%% file: main.tex
\documentclass[10pt,twocolumn,letterpaper]{article}

\usepackage{cvpr}              %

\input{preamble}
\definecolor{cvprblue}{rgb}{0.21,0.49,0.74}
\usepackage[pagebackref,breaklinks,colorlinks,allcolors=cvprblue]{hyperref}
\usepackage{caption}
\usepackage{multirow}
\newcommand{\modelname}{iMFPO}
\def\confName{CVPR}
\def\confYear{2027}

\title{\modelname{}: Improving Mean Flow Policy Optimization with Drift-Aware Velocity Alignment for Embodied Control}

\author{
\textbf{Lisiyu Pan}$^{*}$\quad
\textbf{Ting Huang}$^{*}$\quad
\textbf{Zeyu Zhang}$^{\dag}$\quad
\textbf{Hao Tang}$^{\ddag}$ \\
School of Computer Science, Peking University \\
\small $^*$Equal contribution. $^\dag$Project lead. $^\ddag$Corresponding author: bjdxtanghao@gmail.com
}

\begin{document}
\makeatletter
\let\@oldmaketitle\@maketitle%
\renewcommand{\@maketitle}{\@oldmaketitle%

{\centering
\vspace{-0.35cm}
\includegraphics[width=\linewidth]{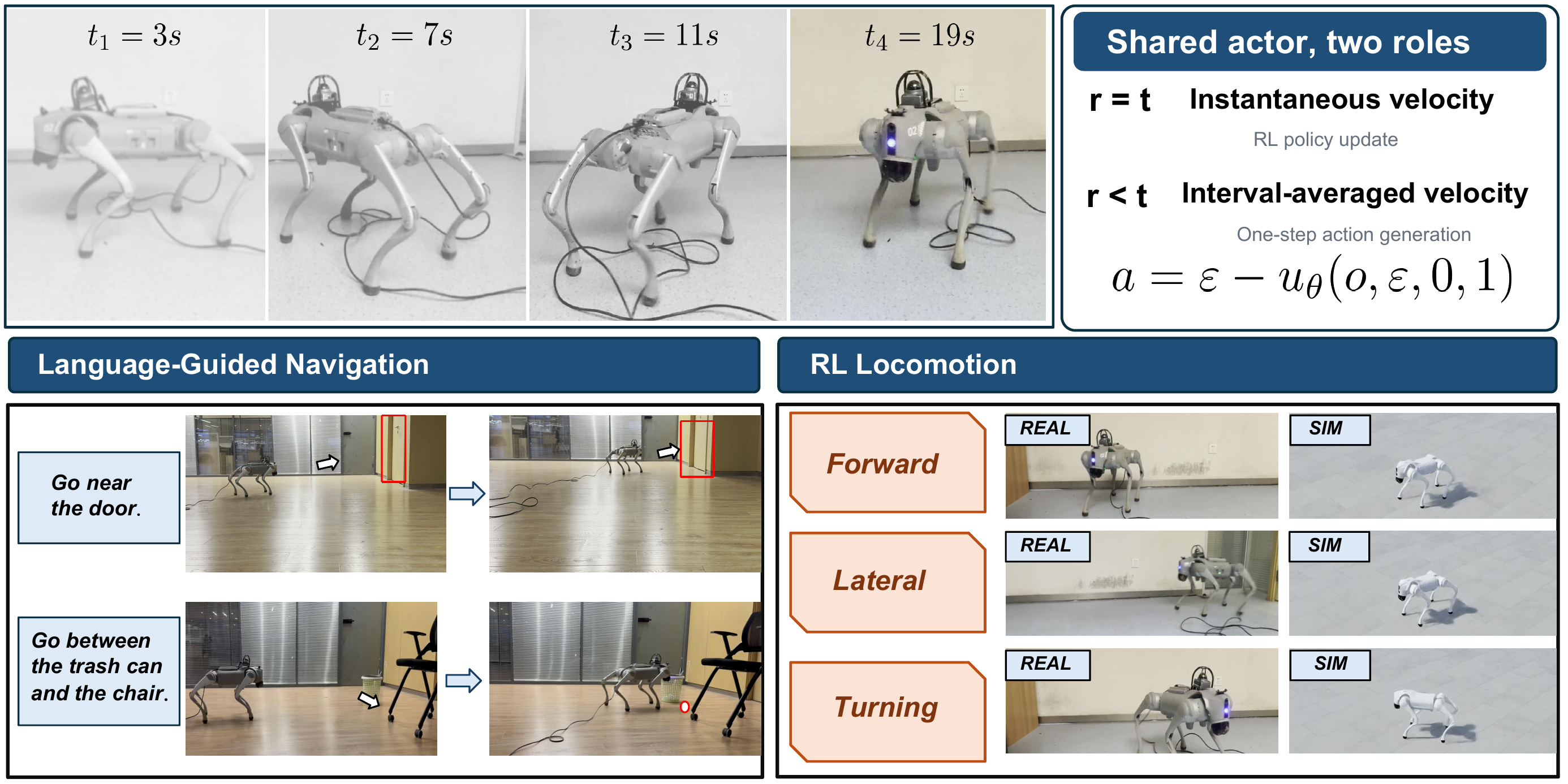}\par
\captionof{figure}{\textbf{\modelname{} for one-step Go2 control.} Real-robot locomotion frames, two language-guided navigation prompts, and real/simulation control examples accompany the shared actor's instantaneous- and interval-velocity roles.
}
\label{fig:teaser}}
\bigskip}%
\makeatother

\maketitle

\begin{abstract}
Flow-based generative policies offer expressive action modeling for embodied control, yet achieving efficient one-step generation under reward-driven policy optimization remains challenging.
In particular, evolving action distributions during online reinforcement learning may introduce inconsistencies between instantaneous-velocity learning and interval-averaged action generation.
To address this challenge, we propose \textbf{\modelname{}}, a one-step generative policy optimization framework with decoupled velocity learning and drift-aware velocity alignment.
A shared interval-conditioned actor supports instantaneous-velocity supervision and interval-averaged action generation.
JVP-corrected velocity regression is jointly optimized with reward-driven policy objectives, with action-level guidance from a frozen PPO teacher during training.
At inference, \modelname{} directly generates actions through interval-averaged velocity prediction, achieving one-step control without iterative flow integration.
Across Go2, G1, H1, and Spot, zero-latent evaluation yields mean returns of $35.69$, $34.44$, $416.23$, and $29.86$.
On Go2, action-generation latency drops from $15.57$\,ms to $0.41$\,ms relative to 64-step RFPO.
\href{https://github.com/AIGeeksGroup/iMFPO}{Code} \&
\href{https://aigeeksgroup.github.io/iMFPO}{Website}.
\end{abstract}

\section{Introduction}

Recent advances in spatial understanding, video reasoning, and vision-language-action learning have expanded the capabilities of embodied intelligence~\cite{huang20253d,huang2026consispace,huang2026mobilevla}.
However, reliable physical interaction requires expressive control policies with low inference latency.
Generative policies offer flexible modeling of complex action distributions.
Diffusion models generate actions through iterative denoising~\cite{ho2020denoising,song2021denoising} and have shown promising results in robotic manipulation and control~\cite{chi2025diffusion,wang2022diffusion,ren2025diffusion}.
Flow matching~\cite{lipman2022flow,liu2022flow} provides an alternative through continuous velocity fields.
Recent flow-based policy optimization methods, including FPO~\cite{mcallister2026flow}, FPO++~\cite{yi2026flow}, and RFPO~\cite{huang2026rfpo}, further improve generative policies through reinforcement learning.
Nevertheless, iterative generation introduces substantial inference overhead, limiting its use in real-time robotic control.

One-step generative modeling offers a promising solution to this limitation.
MeanFlow~\cite{geng2025mean} enables direct generation through interval-averaged velocity prediction, while Improved MeanFlow (iMF)~\cite{geng2026improved} reformulates training as instantaneous-velocity regression.
Recent robotic methods extend these ideas through velocity constraints, directional alignment, and reinforcement learning~\cite{zhan2026mean,fang2025omp,dong2026mean,zou2026one,kim2026score}.
However, integrating one-step velocity learning with online policy optimization remains challenging.
MeanFlow-based objectives involve model-dependent velocity predictions and Jacobian--vector products (JVPs).
Meanwhile, reward-driven policy updates continuously change the action distribution and its associated velocity targets.
This interaction can introduce inconsistencies between instantaneous-velocity supervision and interval-averaged action generation.
A key challenge is therefore to coordinate velocity learning with policy improvement under evolving rollout distributions.

To address this challenge, we propose \textbf{\modelname{}}, a one-step generative policy optimization framework for embodied control (Fig.~\ref{fig:teaser}).
Our framework builds on a shared interval-conditioned actor that supports two complementary velocity representations.
At zero-length intervals, the actor estimates instantaneous velocities for training.
Over the full interval, it predicts the average velocity required for direct noise-to-action generation.
This design connects local velocity supervision with one-step action prediction without introducing separate networks.
It also allows both learning objectives to operate within a unified policy representation.

We further introduce \textbf{drift-aware velocity alignment} to keep instantaneous-velocity supervision, interval-averaged generation, and reward-driven updates consistent as the rollout distribution drifts.
Specifically, JVP-corrected velocity regression provides supervision for the shared velocity field, while a loss-induced policy surrogate aligns that field with reward feedback from online rollouts.
A frozen PPO teacher supplies additional action-level guidance during training.
These components jointly align the actor's velocity predictions under evolving policy distributions.
At inference, the learned policy generates actions with a single network evaluation, without iterative flow integration or teacher guidance.
Experiments across four robotic embodiments demonstrate competitive control performance with substantially reduced inference latency.

The main contributions are summarized as follows:
\begin{itemize}
    \item We propose \textbf{\modelname{}}, a one-step MeanFlow policy optimization framework that decouples instantaneous-velocity learning from interval-averaged action generation through a shared interval-conditioned actor.

    \item We introduce \textbf{drift-aware velocity alignment}, which couples JVP-corrected iMF regression with a velocity-based policy ratio so that reward updates reshape the same interval-conditioned field used for one-step generation under evolving rollouts.

    \item We demonstrate \textbf{competitive one-step control performance} across four robotic embodiments, achieving mean returns of $35.69$, $34.44$, $416.23$, and $29.86$ on Go2, G1, H1, and Spot, respectively. On Go2, \modelname{} reduces action-generation latency from $15.57$\,ms to $0.41$\,ms compared with 64-step RFPO, with additional real-world locomotion demonstrations.
\end{itemize}
\section{Related Work}
\noindent\textbf{Generative policies for robot control.}
Generative policies have emerged as expressive alternatives to conventional unimodal action distributions, enabling the modeling of complex and multimodal robot behaviors.
Diffusion Policy~\cite{chi2025diffusion} demonstrates the effectiveness of diffusion models for visuomotor control, while Diffusion Q-Learning~\cite{wang2022diffusion} and Diffusion Policy Policy Optimization (DPPO)~\cite{ren2025diffusion} extend diffusion-based policies to reinforcement learning (RL).
Alternatively, flow matching~\cite{lipman2022flow} and rectified flow~\cite{liu2022flow} formulate generation through continuous velocity fields, offering efficient transport-based generative modeling.
Recent approaches, including consistency models~\cite{song2023consistency}, FlowPolicy~\cite{zhang2025flowpolicy}, and Flow Q-Learning~\cite{park2025flow}, further explore efficient action generation.
For language-guided legged navigation, NaVILA~\cite{cheng2025navila} combines a VILA-based visual-language model~\cite{lin2024vila} with a locomotion policy through metric mid-level commands.
Real-time robotic control places action quality, optimization efficiency, and inference latency under a shared deployment constraint.

\noindent\textbf{Legged locomotion and visual navigation.}
Early sim-to-real studies demonstrated learned dynamic motor skills~\cite{hwangbo2019learning} and proprioceptive locomotion over challenging terrain~\cite{lee2020learning}.
Massively parallel reinforcement learning enables rapid training of quadrupedal locomotion policies~\cite{rudin2022learning}, while RMA~\cite{kumar2021rma} adapts a simulated policy online to changing terrain and robot dynamics.
Rapid Locomotion~\cite{margolis2022rapid} studies high-speed velocity tracking and transfer to a physical quadruped, and perceptive locomotion~\cite{miki2022learning} uses exteroceptive sensing to traverse challenging terrain.
For vision-and-language navigation, NaVid~\cite{zhang2024navid} predicts the next action from an instruction and a video stream.
Our experiments focus on one-step low-level action generation under velocity commands, including commands issued by NaVILA.

\noindent\textbf{Reinforcement learning for flow policies.}
Building upon proximal policy optimization (PPO)~\cite{schulman2017proximal}, recent studies have investigated RL optimization of flow-based policies.
ReinFlow~\cite{zhang2025reinflow} introduces stochastic transitions into flow trajectories to enable likelihood-based online policy optimization.
Flow Policy Optimization (FPO)~\cite{mcallister2026flow} instead constructs a policy optimization surrogate using conditional flow-matching losses, avoiding explicit likelihood computation.
FPO++~\cite{yi2026flow} further improves optimization stability through per-sample importance ratios and an asymmetric trust region.
To address performance degradation under coarse numerical integration, Rectified Flow Policy Optimization~\cite{huang2026rfpo} employs reward-aware online reflow and teacher-guided supervision for reliable few-step execution.
Refinement-Based Flow Policy Optimization~\cite{park2026refinement} alternatively combines Q-guided action refinement with self-target flow matching to improve the learned action distribution.
Our setting focuses on learning a one-step velocity representation during reward-driven policy updates.

\noindent\textbf{MeanFlow and one-step policy optimization.}
MeanFlow~\cite{geng2025mean} enables one-step generation by directly predicting interval-averaged velocities, eliminating the need for iterative ODE integration.
Improved MeanFlow (iMF)~\cite{geng2026improved} reformulates its self-dependent training objective as instantaneous-velocity regression, while further analysis~\cite{kim2026understanding} investigates the relationship between instantaneous and average velocities.
For robotic control, Mean Velocity Policy (MVP)~\cite{zhan2026mean} introduces instantaneous-velocity boundary constraints, whereas One-step MeanFlow Policy (OMP)~\cite{fang2025omp} employs directional velocity alignment and efficient Jacobian--vector product approximation.
Recent studies have also extended MeanFlow to RL.
Mean Flow Policy Optimization (MFPO)~\cite{dong2026mean} develops a maximum-entropy policy optimization framework, while Dispersive MeanFlow Policy Optimization (DMPO)~\cite{zou2026one} and Score-Based One-step MeanFlow Policy Optimization (SOM)~\cite{kim2026score} incorporate dispersive regularization and Q-guided velocity targets, respectively.
One-Step Flow Q-Learning~\cite{nguyen2026one} further explores average-velocity representations for offline RL.
Our work studies the interaction between instantaneous-velocity supervision, interval-averaged action generation, and reward-driven policy updates within a shared one-step actor.
\begin{figure*}[!t]
    \centering
    \includegraphics[width=\textwidth]{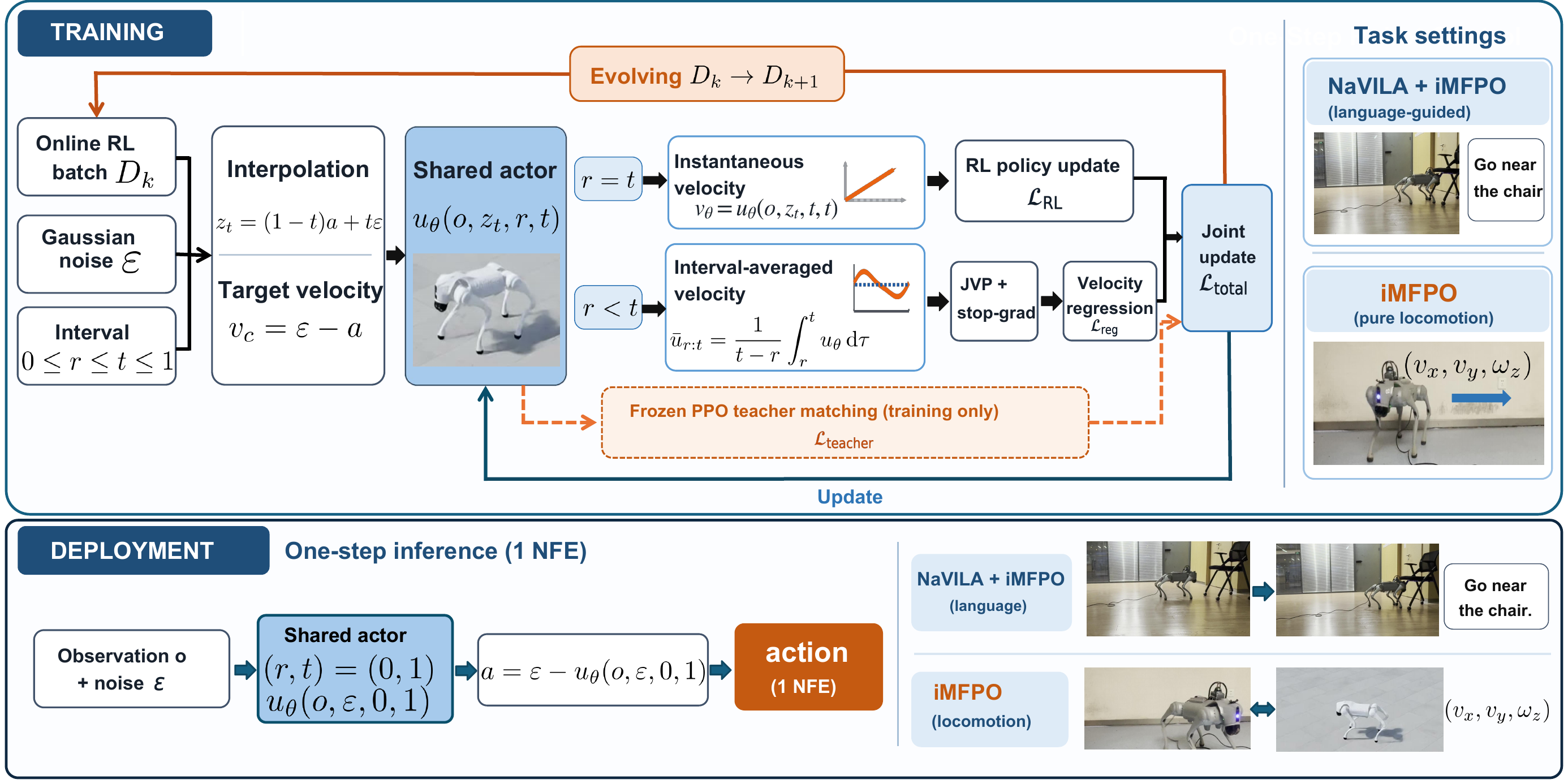}
    \caption{\textbf{Overview of \modelname{}.}
    The shared actor supports instantaneous-velocity learning and one-step action generation.
    Training combines JVP-corrected velocity regression, reward-driven policy updates, and frozen PPO teacher guidance.}
    \label{fig:pipeline}
    \vspace{-0.6cm}
\end{figure*}

\section{Method}
\label{sec:method}

\subsection{Overview}
As illustrated in Fig.~\ref{fig:pipeline}, we present \modelname{}, a one-step generative policy optimization framework that connects local velocity learning with finite-interval action generation through a shared actor.
\emph{Decoupled velocity learning} assigns complementary roles to instantaneous and interval-averaged velocities.
\emph{Drift-aware velocity alignment} couples iMF velocity regression with reward-driven updates on the shared interval-conditioned field under evolving rollout distributions.
During training, auxiliary action-level guidance further regularizes the shared velocity field, promoting stable one-step action generation.
The resulting actor produces actions in a single forward pass without iterative flow integration.

\subsection{Problem Formulation}
We consider an observation-conditioned generative policy that maps a robot observation $o\in\mathcal O$ and an initial latent $\epsilon$ to a continuous action $a\in\mathcal A$.
At policy iteration $k$, the agent collects an on-policy batch $\mathcal D_k$ containing observations, executed actions, and reward-derived learning signals.
Our objective is to improve control performance under the evolving rollout distribution while retaining direct one-step generation.

We parameterize the flow from noise at $t=1$ to actions at $t=0$.
For an action $a$ and Gaussian noise $\epsilon\sim\mathcal N(0,I)$, the interpolated state and its conditional velocity are
\begin{equation}
    z_t=(1-t)a+t\epsilon,\qquad v_c=\epsilon-a,
    \label{eq:imfpo_interpolation}
\end{equation}
where $t\in[0,1]$.
We introduce a shared interval-conditioned actor $u_\theta(o,z_t,r,t)$, defined for $0\leq r\leq t\leq1$, to predict the average velocity between the two time endpoints.
It specifies the finite-interval transport
\begin{equation}
    \hat z_r=z_t-(t-r)u_\theta(o,z_t,r,t).
    \label{eq:imfpo_transport}
\end{equation}
The zero-length interval $r=t$ yields a local velocity estimate, whereas the full interval $(r,t)=(0,1)$ directly maps noise to an action.

\subsection{Decoupled Velocity Learning}

\noindent\textbf{Instantaneous and interval-averaged velocities.}
Consider a probability-flow trajectory $\{z_\tau\}$ with instantaneous velocity $v^\star(o,z_\tau,\tau)$.
Its average velocity over $[r,t]$ is
\begin{equation}
    u^\star(o,z_t,r,t)
    =\frac{1}{t-r}\int_r^t
    v^\star(o,z_\tau,\tau)\,d\tau,\quad r<t.
    \label{eq:imfpo_average}
\end{equation}
Differentiating $(t-r)u^\star=z_t-z_r$ along the trajectory while keeping $r$ fixed gives the MeanFlow identity~\cite{geng2025mean}:
\begin{equation}
    v^\star=u^\star+(t-r)
    \left(\partial_tu^\star+\nabla_z u^\star\,v^\star\right).
    \label{eq:imfpo_identity}
\end{equation}
This relation connects local velocity to finite-interval transport, allowing the two quantities to be represented by one interval-conditioned field.

We exploit the interval endpoints to assign distinct roles to the shared actor.
At $r=t$, it yields the instantaneous velocity
\begin{equation}
    v_\theta(o,z_t,t)\triangleq u_\theta(o,z_t,t,t),
    \label{eq:imfpo_instantaneous}
\end{equation}
which supplies local velocity supervision and the direction for derivative estimation.
For $r<t$, the actor predicts the interval-averaged velocity needed for finite-time transport.
The two roles share parameters, so velocity learning and deployed action prediction are optimized within the same representation rather than with separate networks.

\noindent\textbf{JVP-corrected velocity regression.}
We couple the two velocity regimes through the total-derivative term in Eq.~\eqref{eq:imfpo_identity}.
Using the instantaneous prediction as the tangent direction, the iMF-style Jacobian--vector product (JVP)~\cite{geng2026improved} and corrected velocity are
\begin{equation}
\begin{aligned}
    d_\theta
    &=\operatorname{JVP}_{(z_t,r,t)}
      \!\left[u_\theta;(\operatorname{sg}(v_\theta),0,1)\right],\\
    \tilde v_\theta
    &=u_\theta(o,z_t,r,t)+(t-r)\operatorname{sg}(d_\theta),
\end{aligned}
\label{eq:imfpo_jvp}
\end{equation}
where $o$ and $r$ are fixed during differentiation and $\operatorname{sg}(\cdot)$ denotes stop-gradient.
Detaching the JVP correction prevents gradients from propagating through derivative computation while retaining gradients through the direct actor prediction.
We define the per-sample regression error as the squared residual scaled by the square root of the action dimension:
\begin{equation}
    \ell_{\mathrm{iMF}}(\theta;\xi)
    =\frac{1}{\sqrt{d_a}}
    \left\|\tilde v_\theta-v_c\right\|_2^2,
    \label{eq:imfpo_regression}
\end{equation}
where $d_a$ is the action dimension and $\xi=(o,a,\epsilon,r,t)$.
During training, $r$ is sampled from $\mathcal U(0,t)$, with $r=t$ enforced with probability $p_b$.
Boundary samples reduce Eq.~\eqref{eq:imfpo_regression} to instantaneous-velocity regression, while non-boundary samples supervise the interval-conditioned field.
This shared loss connects local velocity estimation with full-interval action generation.

\subsection{Drift-Aware Velocity Alignment}
Online policy improvement changes the action distribution between rollout iterations, so the actions and velocity targets used to train $u_\theta$ evolve with the policy.
We align reward-driven updates with velocity learning on the current rollout distribution through a shared surrogate.
Instead of introducing a separate action likelihood, we construct the policy ratio from changes in the iMF residual on stored interval-conditioned tuples.
This design keeps reward updates inside the same velocity field that produces one-step actions at deployment, which is the core of drift-aware velocity alignment.
In particular, policy improvement and iMF regression share the stored $(o,a,\epsilon,r,t)$ tuples collected online, so distribution shift is handled without switching the deployed generator.

\noindent\textbf{Velocity-based policy ratio.}
We construct a loss-induced surrogate ratio from the change in iMF regression error.
Let $\theta_k$ denote the behavior policy that collected $\mathcal D_k$.
For a stored tuple $\xi_{ij}=(o_i,a_i,\epsilon_{ij},r_{ij},t_{ij})$, we define
\begin{equation}
    \rho^{\mathrm F}_{ij}(\theta)
    =\exp\!\left[
    \ell_{\mathrm{iMF}}(\theta_k;\xi_{ij})
    -\ell_{\mathrm{iMF}}(\theta;\xi_{ij})\right].
    \label{eq:imfpo_fpo_ratio}
\end{equation}
Both regression errors are evaluated on the same stored tuple, with a clipped log-ratio for numerical stability.
A decrease in $\ell_{\mathrm{iMF}}$ increases $\rho^{\mathrm F}$, so advantageous updates preferentially reshape the interval-conditioned actor toward lower residual on reward-favored samples.
Because each tuple stores the interval endpoints $(r,t)$ used during collection, the same surrogate can act on instantaneous endpoints and longer chords without changing the actor architecture.
As a result, instantaneous supervision through JVP--iMF and full-interval generation remain coupled under one trust-region signal while the rollout distribution drifts.
Reward feedback therefore updates the shared velocity field through the regression-induced policy surrogate rather than through a detached action head.

\noindent\textbf{Constrained policy optimization.}
We optimize $\rho^{\mathrm F}$ with an asymmetric trust-region objective.
Let $\rho=\rho^{\mathrm F}$ and $\bar\rho=\operatorname{clip}(\rho,1-\delta,1+\delta)$ for threshold $\delta>0$.
The per-sample surrogate is
\begin{equation}
    \mathcal S(\rho,\hat A)=
    \begin{cases}
      \max(-\rho\hat A,-\bar\rho\hat A), & \hat A>0,\\[2pt]
      -\rho\hat A+\dfrac{|\hat A|}{2\delta}(\rho-1)^2,
      & \hat A\leq0,
    \end{cases}
    \label{eq:imfpo_aspo}
\end{equation}
with $\mathcal L_{\mathrm{RL}}=\mathbb E[\mathcal S(\rho,\hat A)]$.
The positive-advantage branch employs clipped policy updates, while the non-positive branch imposes a quadratic penalty to constrain deviations from the behavior policy.
This asymmetric treatment limits large destructive updates while still allowing informative positive-advantage steps to reshape $u_\theta$.
Together with $\rho^{\mathrm F}$, the trust region therefore stabilizes online improvement of the one-step velocity field under distribution shift.

\noindent\textbf{Joint Policy Optimization and Action-Level Regularization.}
To further stabilize one-step policy learning, we incorporate auxiliary action-level supervision from a frozen reference policy during training.
Let
$g_\theta(o,\epsilon)=\epsilon-u_\theta(o,\epsilon,0,1)$
denote the one-step prediction of the shared actor.
Denoting the reference policy output by $a_T(o)$, the student is supervised through
\begin{equation}
    \mathcal L_T
    =\mathbb E_{o,\tilde\epsilon}
    \!\left[\frac{1}{d_a}\left\|
    g_\theta(o,\tilde\epsilon)
    -\operatorname{sg}(a_T(o))\right\|_2^2\right],
    \label{eq:imfpo_teacher}
\end{equation}
where $\tilde\epsilon$ follows the training mixture of Gaussian and zero initializations.
Teacher matching regularizes the deployed one-step map without changing the inference graph.
The overall optimization objective is
\begin{equation}
    \mathcal L_{\mathrm{total}}
    =\lambda_\pi\mathcal L_{\mathrm{RL}}
    +\lambda_V\mathcal L_V
    +\lambda_T\mathcal L_T,
    \label{eq:imfpo_total}
\end{equation}
where $\mathcal L_V$ is the critic regression loss.
Our unified optimization framework integrates the velocity-based policy surrogate with auxiliary action-level regularization, jointly aligning reward-driven policy improvement and interval-conditioned velocity learning within a shared representation.
This choice matches the deployed full-interval map in Eq.~\eqref{eq:imfpo_action}: reward updates reshape the same velocity field that produces $a_\theta(o,\epsilon)$ at inference.
The resulting policy operates as a self-contained one-step generator, directly producing actions through the learned interval-averaged velocity field without auxiliary inference components.

\subsection{Direct One-Step Action Generation}
At deployment, we evaluate the learned actor on the full interval $(r,t)=(0,1)$.
Given an observation $o$ and initial latent $\epsilon$, the generated action is
\begin{equation}
    a_\theta(o,\epsilon)
    =\epsilon-u_\theta(o,\epsilon,0,1).
    \label{eq:imfpo_action}
\end{equation}
The latent can be Gaussian noise $\epsilon\sim\mathcal N(0,I)$ or a zero vector $\epsilon=0$, according to the evaluation protocol.
Since the full-interval displacement is predicted directly, action generation requires one network function evaluation (1 NFE), with neither numerical flow integration nor teacher inference.
\section{Experiments}
\begin{table}[!t]
\centering
\caption{\textbf{Locomotion performance across four embodiments.}
Mean episode return; higher is better.}
\label{tab:imfpo_four_robots}
\small
\setlength{\tabcolsep}{3pt}
\renewcommand{\arraystretch}{1.08}
\begin{tabular*}{\linewidth}{@{\extracolsep{\fill}}lccccc@{}}
\toprule
Method & NFE & Go2 & G1 & H1 & Spot \\
\midrule
\multicolumn{6}{l}{\emph{Zero latent}} \\ \midrule
FPO++~\cite{yi2026flow} & \multirow{2}{*}{64} & $18.56$ & $18.17$ & $380.00$ & $9.96$ \\
RFPO~\cite{huang2026rfpo} & & $35.58$ & $32.55$ & $416.06$ & $28.19$ \\ \midrule
FPO++~\cite{yi2026flow} & \multirow{3}{*}{1} & $-185.70$ & $-1.57$ & $19.38$ & $-462.99$ \\
RFPO~\cite{huang2026rfpo} & & $35.16$ & $33.59$ & $412.40$ & $27.89$ \\
\textbf{\modelname{} (Ours)} & & $\mathbf{35.69}$ & $\mathbf{34.44}$ & $\mathbf{416.23}$ & $\mathbf{29.86}$ \\
\midrule \midrule
\multicolumn{6}{l}{\emph{Random latent}} \\ \midrule
FPO++~\cite{yi2026flow} & \multirow{2}{*}{64} & $17.96$ & $3.96$ & $340.28$ & $9.16$ \\
RFPO~\cite{huang2026rfpo} & & $33.03$ & $8.84$ & $404.32$ & $24.58$ \\ \midrule
FPO++~\cite{yi2026flow} & \multirow{3}{*}{1} & $-211.67$ & $-2.78$ & $23.84$ & $-204.53$ \\
RFPO~\cite{huang2026rfpo} & & $33.30$ & $17.71$ & $397.55$ & $24.27$ \\
\textbf{\modelname{} (Ours)} & &$\mathbf{35.34}$ & $\mathbf{30.79}$ & $\mathbf{406.77}$ & $\mathbf{30.15}$ \\
\bottomrule
\end{tabular*}
\vspace{-0.5cm}
\end{table}

\subsection{Experimental Setup}

\noindent\textbf{Tasks and embodiments.}
We evaluate \modelname{} on velocity-tracking locomotion in Isaac Lab with a unified velocity-control interface across four embodiments: Go2, G1-29DoF, H1, and Spot.
Each embodiment uses its native robot model and robot-specific observation, action, reward, and termination definitions.
Policies receive proprioceptive observations and commanded planar velocities $(v_x,v_y,\omega_z)$, and output continuous low-level actions.
For a given embodiment, all methods share the same task implementation and inference protocol.
We also study recorded Go2 command replay in Isaac Lab and hierarchical language-guided control in MuJoCo.

\noindent\textbf{Baselines and evaluation protocol.}
We compare with FPO++~\cite{yi2026flow} and RFPO~\cite{huang2026rfpo} at 64 and 1 network function evaluations (NFEs).
\modelname{} uses 1 NFE throughout.
We report mean episode return under \emph{Zero} ($\epsilon=0$) and \emph{Random} ($\epsilon\sim\mathcal N(0,I)$) initial latents.
Unless otherwise specified, each result averages 256 evaluation episodes.

\begin{table}[!t]
\centering
\caption{\textbf{Go2 control performance and inference latency.}
Zero-latent evaluation with 256 episodes.
Latency is batch-size-one GPU actor latency (p50), excluding environment stepping.}
\label{tab:imfpo_go2_efficiency}
\small
\setlength{\tabcolsep}{4pt}
\renewcommand{\arraystretch}{1.12}
\begin{tabular}{lrrr}
\toprule
Method & NFE & Return $\uparrow$ & Latency (ms) $\downarrow$ \\
\midrule
FPO++~\cite{yi2026flow} & 64 & $18.56$ & $14.94$ \\
FPO++ & 1 & $-185.70$ & $1.18$ \\
RFPO~\cite{huang2026rfpo} & 64 & $35.58$ & $15.57$ \\
RFPO & 1 & $35.16$ & $0.39$ \\
MFPO~\cite{dong2026mean} & 1 & $35.54$ & $0.41$ \\
\textbf{\modelname{} (Ours)} & 1 & $\mathbf{35.69}$ & $0.41$ \\
\bottomrule
\end{tabular}
\vspace{-0.6cm}
\end{table}

\noindent\textbf{Implementation details.}
The actor is a three-layer MLP with 256 hidden units per layer.
The critic is a three-layer MLP with 768 hidden units per layer.
Both use ELU activations.
We optimize with AdamW (learning rate $1{\times}10^{-4}$, weight decay $1{\times}10^{-4}$).
Training uses $4096$ parallel environments on Go2, H1, and Spot, and $2048$ on G1.
The PPO teacher distillation weight is $0.1$ in the first stage and $0.2$ in the polish stage.
The interval endpoint is sampled as $r\sim\mathcal U(0,t)$ with boundary probability $p_b=0.25$.
All four embodiments use the $\rho^{\mathrm F}$ recipe with frozen PPO teacher matching at training time, and deploy a single-step actor at 1 NFE.

\subsection{Comparative Analysis}

\noindent\textbf{Performance across robotic embodiments.}
Tab.~\ref{tab:imfpo_four_robots} summarizes returns across four embodiments under a matched evaluation protocol.
Under zero-latent initialization, \modelname{} obtains $35.69$, $34.44$, $416.23$, and $29.86$ on Go2, G1, H1, and Spot with a single network evaluation.
These scores are comparable to, or slightly above, the corresponding 64-NFE RFPO baselines ($35.58$, $32.55$, $416.06$, and $28.19$).
This indicates that a policy trained for full-interval MeanFlow prediction can match a strong multi-step flow optimizer at test time with one network evaluation.
The cross-embodiment consistency further indicates that the shared one-step actor transfers across quadruped and humanoid morphologies under the same evaluation protocol.
Reducing FPO++ from 64 NFEs to 1 NFE causes large degradations on every robot (for example, Go2 falls from $18.56$ to $-185.70$).
This pattern indicates that a multi-step solver trained for fine numerical integration becomes unreliable when forced to act with a single coarse step.
\modelname{} instead optimizes the deployed map in Eq.~\eqref{eq:imfpo_action} during training, so reward updates reshape the same interval-conditioned field used at inference.
Aligning the training objective with the inference map therefore explains why one-step deployment remains competitive with 64-NFE RFPO.

Under random latents, \modelname{} remains stable on Go2 ($35.34$), G1 ($30.79$), H1 ($406.77$), and Spot ($30.15$).
1-NFE RFPO stays close to its 64-NFE score on Go2 and Spot, but drops sharply on G1 from $33.59$ (Zero) to $17.71$ (Random).
The G1 gap indicates that a $\rho^{\mathrm F}$-trained full-interval policy keeps the noise endpoint inside the transport used online.
The same Random protocol also shows that \modelname{} preserves near-zero gaps on Go2, H1, and Spot, which supports Contribution~2 beyond a single embodiment.

\noindent\textbf{One-step inference efficiency.}
Tab.~\ref{tab:imfpo_go2_efficiency} isolates the return--latency trade-off on Go2.
\modelname{} reaches $35.69$ return at $0.41$\,ms batch-size-one actor latency.
64-NFE RFPO reaches a similar return ($35.58$) at $15.57$\,ms, an approximately $38\times$ reduction in measured action-generation latency.
This latency gap matches the Method claim that full-interval MeanFlow generation removes iterative ODE integration at deployment.
Under random latents on Go2, \modelname{} scores $35.34$, while 64-NFE RFPO scores $33.03$.
Reading return and latency together indicates that \modelname{} delivers multi-step-level return at 1-NFE latency, with stronger random-latent behavior than RFPO on this benchmark.
MFPO obtains $35.54$ return at the same $0.41$\,ms latency.
FPO++ at 1 NFE collapses to $-185.70$ at $1.18$\,ms, which indicates that low latency is useful only when the policy is trained for that operating point.
Overall, the table links teacher-free one-step inference with a practical wall-clock advantage on Go2.

\begin{table}[!t]
\centering
\caption{\textbf{Ablation study on Go2.}
Mean episode return under zero and random initial latents.}
\label{tab:imfpo_ablation}
\small
\setlength{\tabcolsep}{3.5pt}
\renewcommand{\arraystretch}{1.10}
\begin{tabular}{lcc}
\toprule
Variant & Zero $\uparrow$ & Random $\uparrow$ \\
\midrule
Full \modelname{} & $\mathbf{35.69}$ & $\mathbf{35.34}$ \\
W/o JVP--iMF & $29.21$ & $26.24$ \\
Full-interval sampling ($r=0$) & $34.93$ & $34.88$ \\
W/o interval loss & $21.00$ & $18.91$ \\
\bottomrule
\end{tabular}
\vspace{-0.4cm}
\end{table}

\begin{figure}[t]
    \centering
    \includegraphics[width=\linewidth]{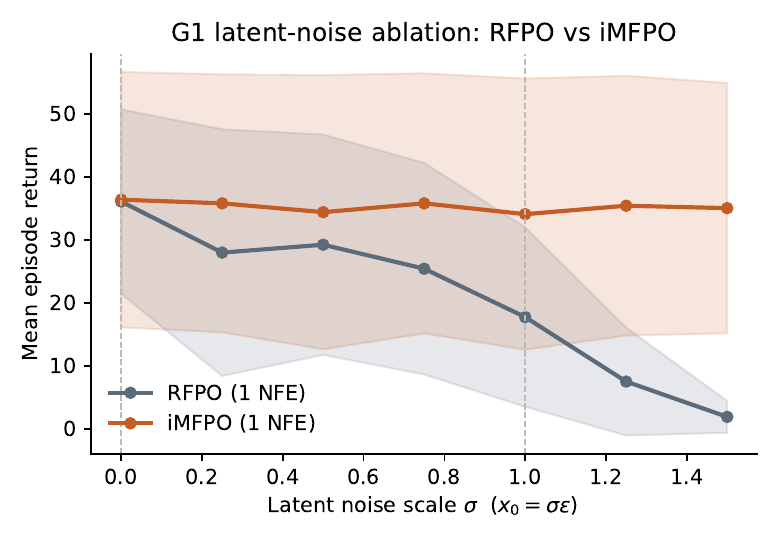}
    \caption{\textbf{Latent-noise robustness on G1.} Mean episode return under different initial noise scales. Each point averages 256 episodes.}
    \label{fig:g1_latent_noise}
    \vspace{-0.6cm}
\end{figure}

\begin{figure*}[!t]
    \centering
    \includegraphics[width=\linewidth]{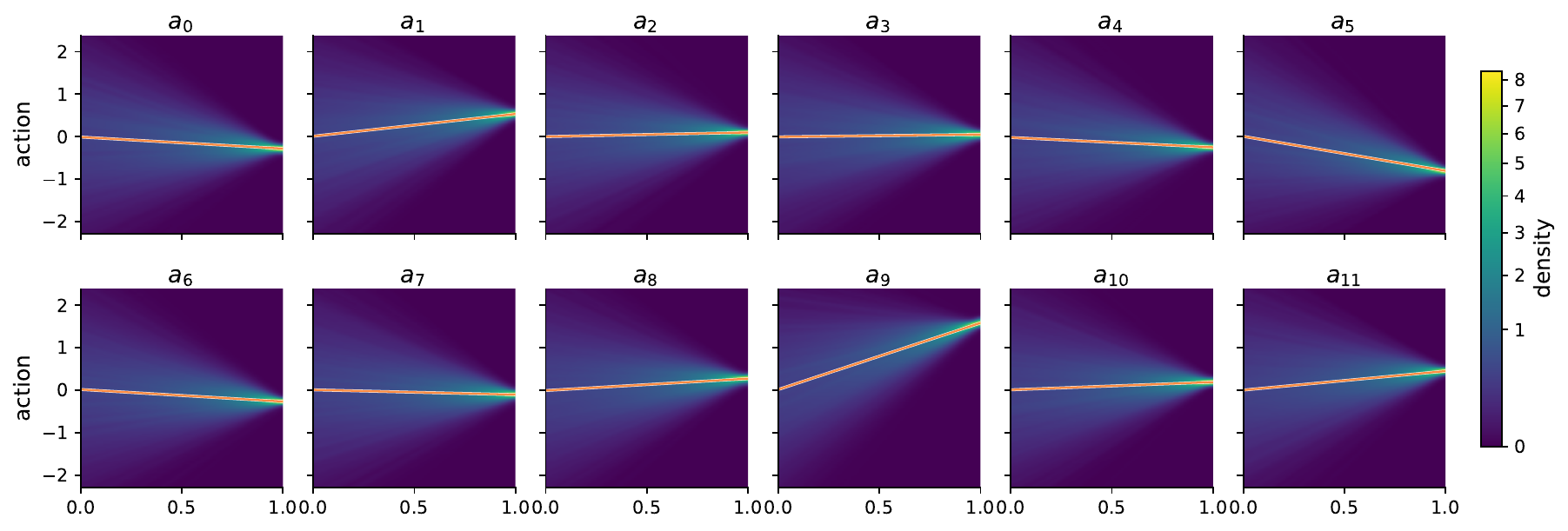}
    \caption{\textbf{Action-flow visualization on Go2.} Marginal densities of 12 action dimensions over $10{,}000$ samples. Horizontal axis is generation progress $s{=}1{-}t$ ($0$: noise, $1$: action), with chords $z_t=(1-t)a+t\epsilon$. Orange curves show mean trajectories.}
    \label{fig:action_flow_geometry}
    \vspace{-0.4cm}
\end{figure*}

\begin{figure*}[!t]
    \centering
    \includegraphics[width=\linewidth]{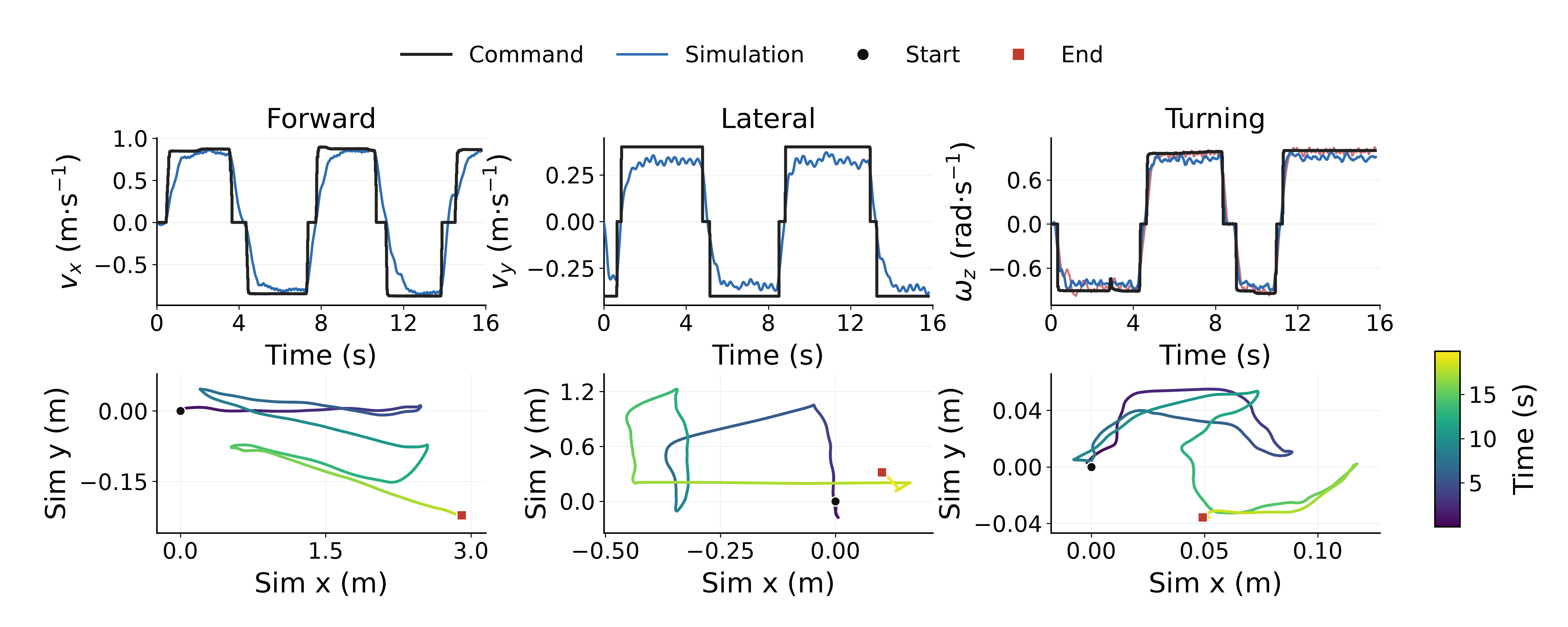}
    \caption{\textbf{Go2 velocity tracking in Isaac Lab.} Commanded and simulated velocities (top) and planar trajectories (bottom). The controller runs at 50 Hz with 1 NFE.}
    \label{fig:real_sim_replay}
    \vspace{-0.6cm}
\end{figure*}
\subsection{Ablation Study}

\noindent\textbf{Component ablation.}
Tab.~\ref{tab:imfpo_ablation} isolates design choices that implement decoupled velocity learning on Go2.
Removing JVP--iMF lowers zero- and random-latent returns from $35.69$ and $35.34$ to $29.21$ and $26.24$.
The larger random-latent drop indicates that local--interval consistency matters more when the initial latent varies, supporting Contribution~1 and the JVP correction in Eq.~\eqref{eq:imfpo_jvp}.
Fixing every sample to the full interval ($r=0$) yields $34.93$ and $34.88$, a smaller gap relative to the full model.
This small change indicates that matching the deployed interval $(r,t)=(0,1)$ already captures most of the one-step objective, while random-$r$ sampling provides an additional refinement.
Removing the interval iMF regression term yields $21.00$ and $18.91$, which indicates that finite-interval velocity learning is essential for stable one-step control under the same $\rho^{\mathrm F}$+KD recipe.
Taken together, JVP--iMF and interval regression are the main drivers of the Full model, while full-interval sampling is a milder change.
Because all variants share the same architecture and training budget, the gaps attribute performance differences to the velocity-learning objective.

\noindent\textbf{Latent-noise robustness.}
Fig.~\ref{fig:g1_latent_noise} provides a controlled stress test of drift-aware velocity alignment on G1.
We evaluate return under $\epsilon\leftarrow\sigma\xi$ with $\xi\sim\mathcal N(0,I)$ for $\sigma\in\{0,0.25,0.5,0.75,1.0,1.25,1.5\}$, where $\sigma=0$ and $\sigma=1$ recover the Zero and Random protocols.
At matched $\sigma=0$ starting points, both methods are competitive, so later divergence is attributable to increasing noise scale.
As $\sigma$ increases, 1-NFE RFPO degrades sharply, whereas $\rho^{\mathrm F}$-trained \modelname{} stays nearly flat.
This indicates that drift-aware velocity alignment keeps moderate noise-endpoint perturbations inside a transport optimized end-to-end by $\rho^{\mathrm F}$ and iMF.
The smooth growth of the gap with $\sigma$ complements the large Random gap on G1 in Tab.~\ref{tab:imfpo_four_robots}.
The sweep therefore provides continuous evidence for Contribution~2 across a range of latent scales rather than a single operating point.

\begin{figure*}[!t]
    \centering
    \includegraphics[width=\linewidth]{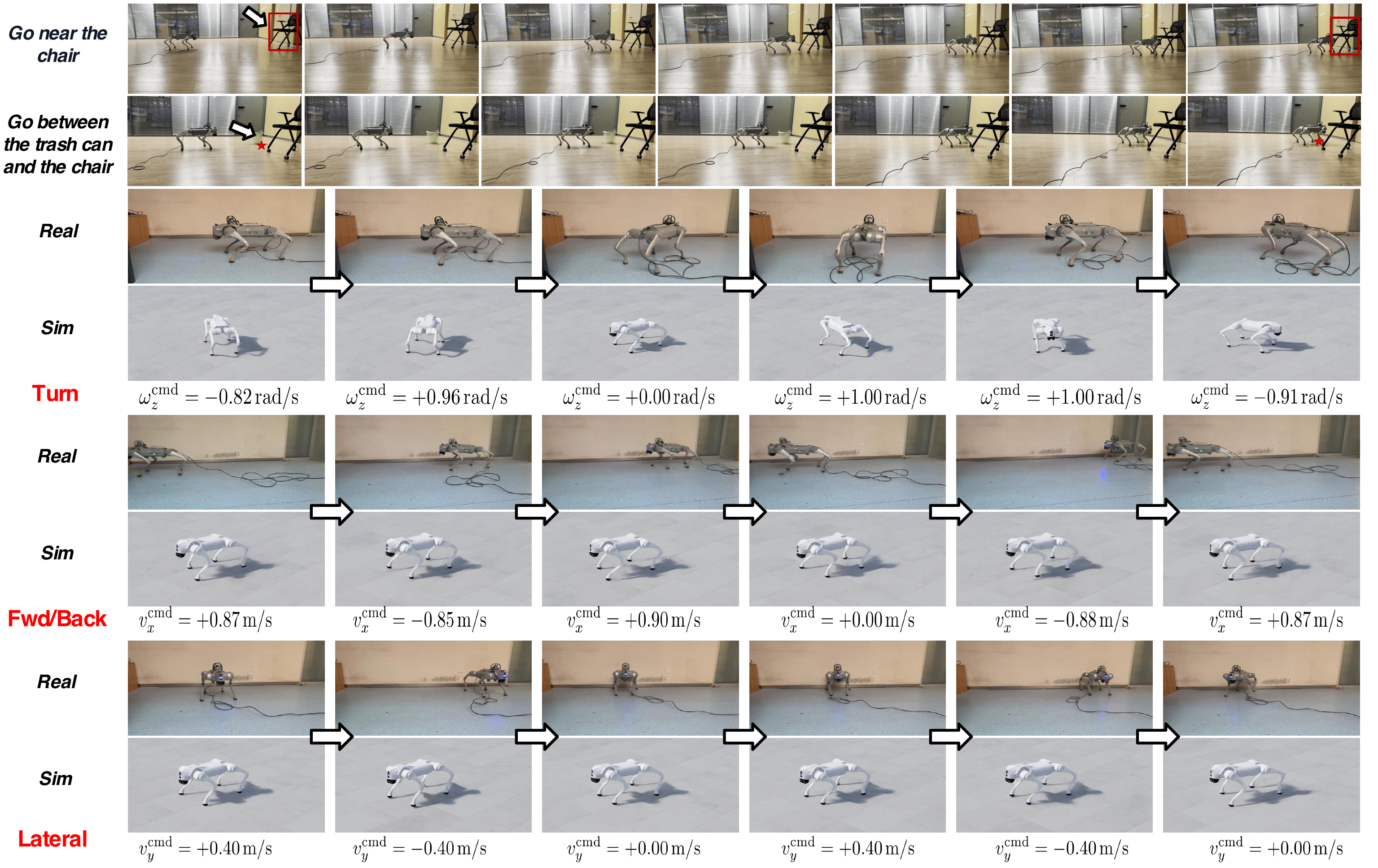}
    \caption{\textbf{Go2 navigation and locomotion keyframes.} Real-world navigation sequences (top) and real/simulated locomotion examples (bottom).}
    \label{fig:real_world_keyframes}
    \vspace{-0.6cm}
\end{figure*}

\subsection{Qualitative Analysis}

\noindent\textbf{Action-flow visualization.}
Fig.~\ref{fig:action_flow_geometry} visualizes the action distributions generated by \modelname{} on Go2.
For a fixed observation, we sample multiple noise inputs and visualize the marginal densities of all twelve action dimensions.
The distributions gradually concentrate from noise to actions, with orange curves indicating the mean trajectories.
The horizontal axis denotes generation progress $s=1-t$, from noise ($s=0$) to actions ($s=1$).

\noindent\textbf{Velocity tracking.}
We replay three recorded Go2 velocity-command trajectories in Isaac Lab using the 1-NFE \modelname{} controller at 50 Hz.
The trajectories cover forward, lateral, and turning motions, each lasting approximately 19--20 seconds.
All simulated rollouts complete without falls.
The mean absolute tracking errors range from $0.014$ to $0.145$\,m/s longitudinally and from $0.025$ to $0.093$\,m/s laterally.
Fig.~\ref{fig:real_sim_replay} shows the velocity responses and planar trajectories.
The results demonstrate stable velocity tracking under different motion commands.

\noindent\textbf{Real-world demonstrations.}
Fig.~\ref{fig:real_world_keyframes} presents real-world Go2 navigation sequences and real/simulation locomotion comparisons.
The results illustrate the deployment of the 1-NFE controller on a physical robot and its integration with high-level navigation commands.

\subsection{Language-Guided Control}

We connect NaVILA~\cite{cheng2025navila} to the frozen Go2 \modelname{} locomotion controller.
Given a language instruction, NaVILA generates mid-level navigation commands, which are converted into planar velocity targets and executed by the 1-NFE controller.
We consider three spatial instructions: ``Go near the chair,'' ``Go between the trash can and the chair,'' and ``Go near the door.''
Figs.~\ref{fig:teaser} and~\ref{fig:real_world_keyframes} show the corresponding execution sequences.
For the chair instruction, we evaluate a MuJoCo scene with an obstacle between the robot and the target.
A scripted mid-level command sequence brings the robot within $0.52$\,m of the chair without a fall.
NaVILA-driven rollouts and the remaining instructions are shown in Figs.~\ref{fig:teaser} and~\ref{fig:real_world_keyframes}.
These results illustrate the integration of language-guided navigation with one-step locomotion control.
\section{Limitations}
Our quantitative evaluation focuses on flat velocity tracking in simulation, without systematic terrain or disturbance stress tests.
Reported returns use selected checkpoints, and real-robot results are qualitative rather than full tracking or navigation success rates.

\section{Conclusion}
We presented \modelname{}, a one-step generative policy optimization framework for embodied control.
A shared actor decouples instantaneous-velocity learning from interval-averaged action generation, and drift-aware velocity alignment couples iMF regression with reward-driven updates under evolving rollouts.
Across four embodiments, \modelname{} supports competitive one-step control.
On Go2, it reaches $35.69$ return while reducing action-generation latency from $15.57$\,ms to $0.41$\,ms versus 64-step RFPO.
Command replay and language-guided control further illustrate the controller in extended pipelines.
Future work will study broader alignment objectives and real-world generalization.

\section*{LLM Usage Statement}
OpenAI ChatGPT was used to assist with language editing and drafting portions of the manuscript.
The authors remain responsible for all scientific claims, citations, and experimental results.

{
    \small
    \bibliographystyle{ieeenat_fullname}
    \bibliography{main}
}

\input{X_suppl}

\end{document}

%% file: X_suppl.tex
\clearpage
\setcounter{page}{1}
\maketitlesupplementary

\setcounter{section}{0}
\renewcommand{\thesection}{\Alph{section}}
\setcounter{table}{0}
\renewcommand{\thetable}{S\arabic{table}}
\setcounter{figure}{0}
\renewcommand{\thefigure}{S\arabic{figure}}

\section{Additional Experimental Details}
\label{sec:supple_hyperparams}

\subsection{Training Configurations}

We provide additional implementation details for the locomotion experiments across Go2, G1, H1, and Spot.
All embodiments share the same interval-conditioned policy representation and one-step action generation formulation, while their policy optimization objectives follow the corresponding exploration configurations.

The actor and critic are implemented as multilayer perceptrons with ELU activations. The actor contains three hidden layers of 256 units, while the critic uses three hidden layers of 768 units.
The shared optimization hyperparameters are summarized in Tab.~\ref{tab:app_shared_config}.

\begin{table}[bhtp]
\vspace{-0.3cm}
\small
\centering
\caption{\textbf{Shared training configurations.}}
\label{tab:app_shared_config}
\small
\setlength{\tabcolsep}{4pt}
\renewcommand{\arraystretch}{1.12}
\begin{tabular}{lc}
\toprule
Setting & Value \\
\midrule
Optimizer & AdamW \\
Learning rate & $1\times10^{-4}$ \\
Discount factor $\gamma$ & $0.99$ \\
GAE parameter $\lambda$ & $0.95$ \\
Optimization epochs & $16$ \\
Mini-batches & $4$ \\
Rollout horizon (steps) & $24$ \\
Trust-region bound $\delta$ & $0.05$ \\
Actor architecture & $[256,256,256]$ \\
Critic architecture & $[768,768,768]$ \\
Activation & ELU \\
Interval sampling & $r\sim\mathcal U(0,t)$ \\
Boundary probability $p_b$ & $0.25$ \\
\bottomrule
\end{tabular}
\end{table}

Tab.~\ref{tab:app_robot_config} summarizes the embodiment-specific environment configurations.
The action dimensionality follows the corresponding robot control interfaces.

\begin{table}[bhtp]
\vspace{-0.3cm}
\small
\centering
\caption{\textbf{Robot-specific configurations.}}
\label{tab:app_robot_config}
\small
\setlength{\tabcolsep}{8pt}
\renewcommand{\arraystretch}{1.12}
\begin{tabular}{lcc}
\toprule
Robot & Parallel environments & Action dimension \\
\midrule
Go2 & $4096$ & $12$ \\
G1 (29-DoF) & $2048$ & $29$ \\
H1 & $4096$ & $19$ \\
Spot & $4096$ & $12$ \\
\bottomrule
\end{tabular}
\end{table}

The Go2 training schedule consists of an initial optimization stage followed by checkpoint-based refinement.
The initial stage runs for 1,500 iterations with a teacher-matching coefficient of $0.1$.
The subsequent refinement stage runs for 1,001 iterations, increasing the coefficient to $0.2$.
The Spot training curve reports the initial 1,500-iteration stage.

\subsection{Policy Optimization and Inference}

\noindent\textbf{Interval-conditioned training.}
At each optimization iteration, an on-policy batch is collected using the current actor.
Given an action $a$ and Gaussian noise $\epsilon$, an intermediate state is constructed as
\begin{equation}
    z_t=(1-t)a+t\epsilon.
\end{equation}
The interval endpoint is sampled according to $r\sim\mathcal U(0,t)$, with the boundary case $r=t$ selected with probability $p_b=0.25$.
Both interval-averaged and instantaneous velocity predictions are obtained from the same actor.

\noindent\textbf{Embodiment-specific policy updates.}
The policy optimization follows the exploration configuration of each embodiment.
For Go2, the iMF regression error contributes to the loss-induced policy ratio used for reward-driven policy improvement.
For G1, H1, and Spot, a diagonal Gaussian exploration policy is used, with the one-step actor prediction serving as its mean.
An additional advantage-weighted iMF regression objective supervises the velocity representation using positive-advantage samples.

\noindent\textbf{Teacher-guided training.}
A frozen PPO teacher provides action-level supervision through a teacher-matching objective.
The teacher parameters remain fixed throughout student optimization.
This supervision is applied only during training and does not introduce additional inference-time computation.

\noindent\textbf{Deployment.}
At deployment, the actor is evaluated over the full interval $(r,t)=(0,1)$:
\begin{equation}
    a_\theta(o,\epsilon)
    =\epsilon-u_\theta(o,\epsilon,0,1).
\end{equation}
Only the shared actor is retained.
The critic, frozen teacher, velocity-regression objective, and JVP computation are discarded.
Consequently, each action is generated using a single network evaluation, without iterative flow integration.

\begin{figure*}[t]
\centering
\includegraphics[width=0.96\textwidth]{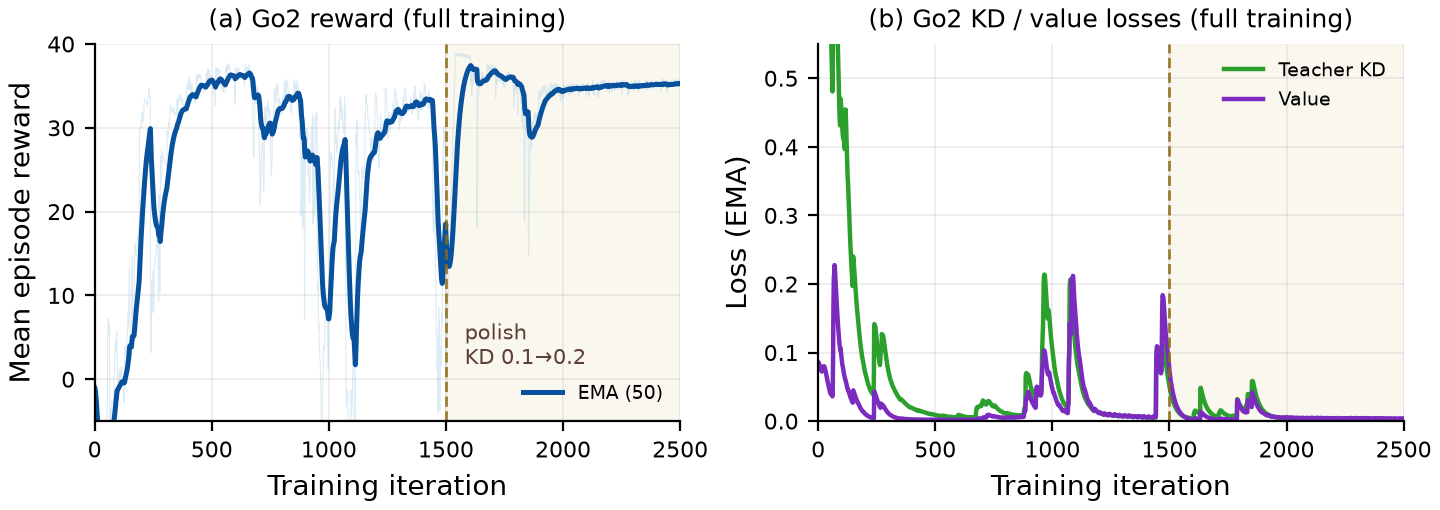}
\caption{\textbf{Training dynamics of iMFPO on Go2.}
(a) Mean episode reward during the initial optimization and refinement stages. The dashed line marks the transition at 1,500 iterations, where the teacher-matching coefficient increases from $0.1$ to $0.2$.
(b) Corresponding teacher-matching and value losses.
}
\label{fig:app_go2_dynamics}
\end{figure*}

\begin{figure*}[t]
\centering
\includegraphics[width=0.96\textwidth]{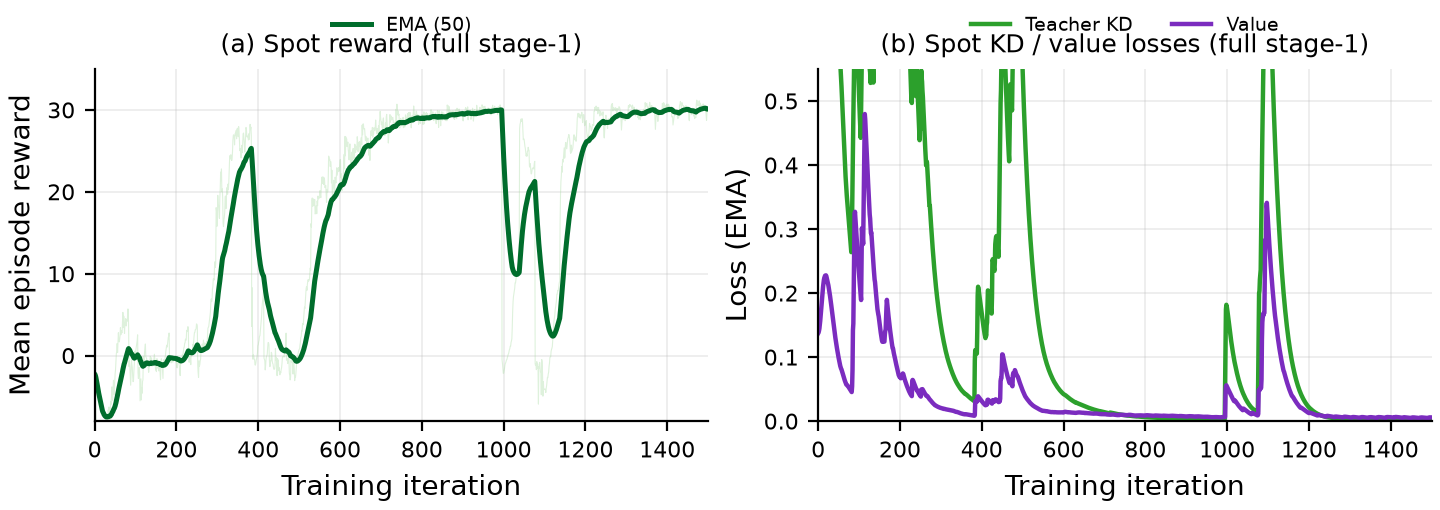}
\caption{\textbf{Training dynamics of iMFPO on Spot.}
(a) Mean episode reward during the initial training stage.
(b) Corresponding teacher-matching and value losses.
}
\label{fig:app_spot_dynamics}
\end{figure*}

\section{Training Dynamics}
\label{sec:app_training_dynamics}

\subsection{Go2 Training Dynamics}

Fig.~\ref{fig:app_go2_dynamics} presents the complete training trajectory of iMFPO on Go2.
The initial optimization stage progressively improves the mean episode reward, although noticeable fluctuations occur throughout online training.
At the transition to checkpoint-based refinement, the teacher-matching coefficient is increased from $0.1$ to $0.2$.
The reward subsequently recovers and approaches approximately $35$ toward the end of training.

The teacher-matching and value losses generally decrease during optimization, with transient peaks at several update intervals.
These curves provide complementary diagnostics of the interaction between policy optimization and auxiliary supervision.

\subsection{Spot Training Dynamics}
\label{sec:app_spot_dynamics}

Fig.~\ref{fig:app_spot_dynamics} shows the training trajectory on Spot during the initial optimization stage.
The mean episode reward gradually increases and reaches approximately $30$.
Although several performance drops occur during training, the reward subsequently recovers.
The teacher-matching and value losses exhibit corresponding transient fluctuations before decreasing toward smaller values.

Together, the Go2 and Spot curves provide additional training diagnostics across two different robotic embodiments, complementing the final evaluation results reported in the main paper.

\section{Additional Qualitative Results}
\label{sec:app_qualitative}

\subsection{Cross-Embodiment Locomotion Visualizations}
\label{sec:app_cross_embodiment}

\begin{figure*}[t]
\centering
\includegraphics[width=\textwidth]{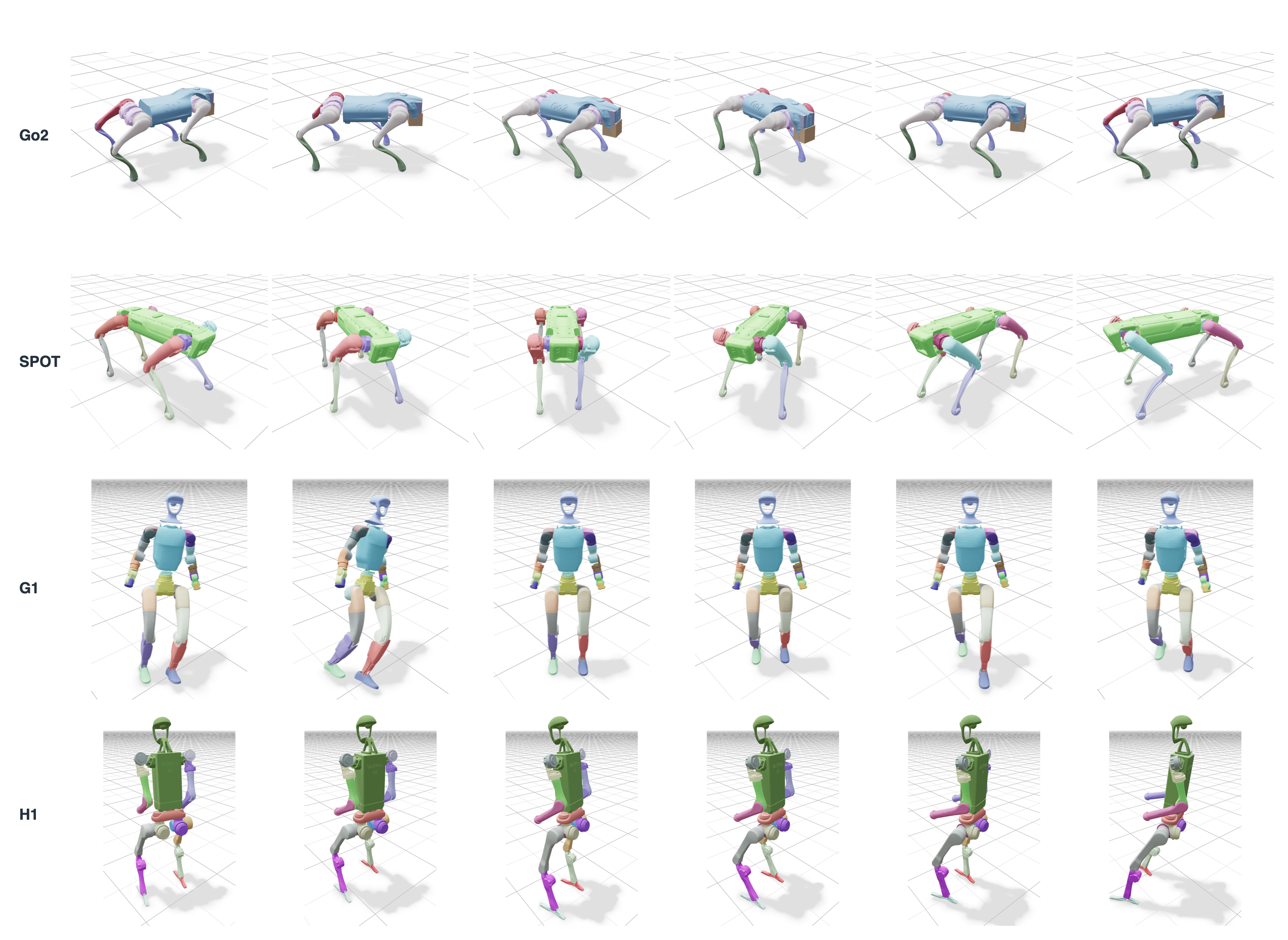}
\caption{\textbf{Cross-embodiment locomotion visualization (I).}
Representative simulation rollout frames of Go2, Spot, G1, and H1 under the learned one-step generative policies.}
\label{fig:app_robot_rollouts_straight}
\vspace{-0.4cm}
\end{figure*}

\begin{figure*}[t]
\centering
\includegraphics[width=\textwidth]{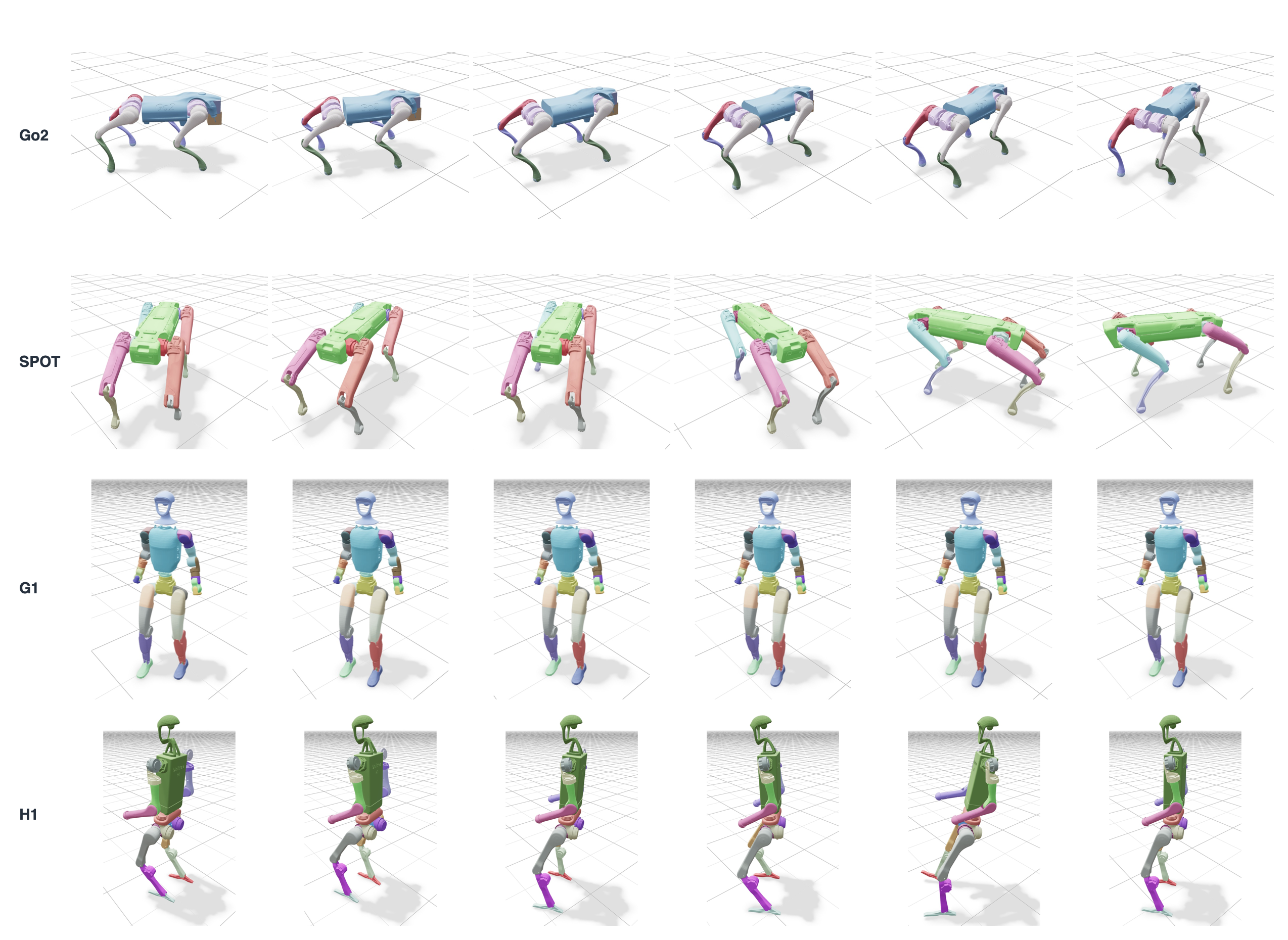}
\caption{\textbf{Cross-embodiment locomotion visualization (II).}
Additional simulation rollout frames illustrating policy execution across the four robotic embodiments.}
\label{fig:app_robot_rollouts_lateral}
\end{figure*}

\begin{figure*}[t]
\centering
\includegraphics[width=\textwidth]{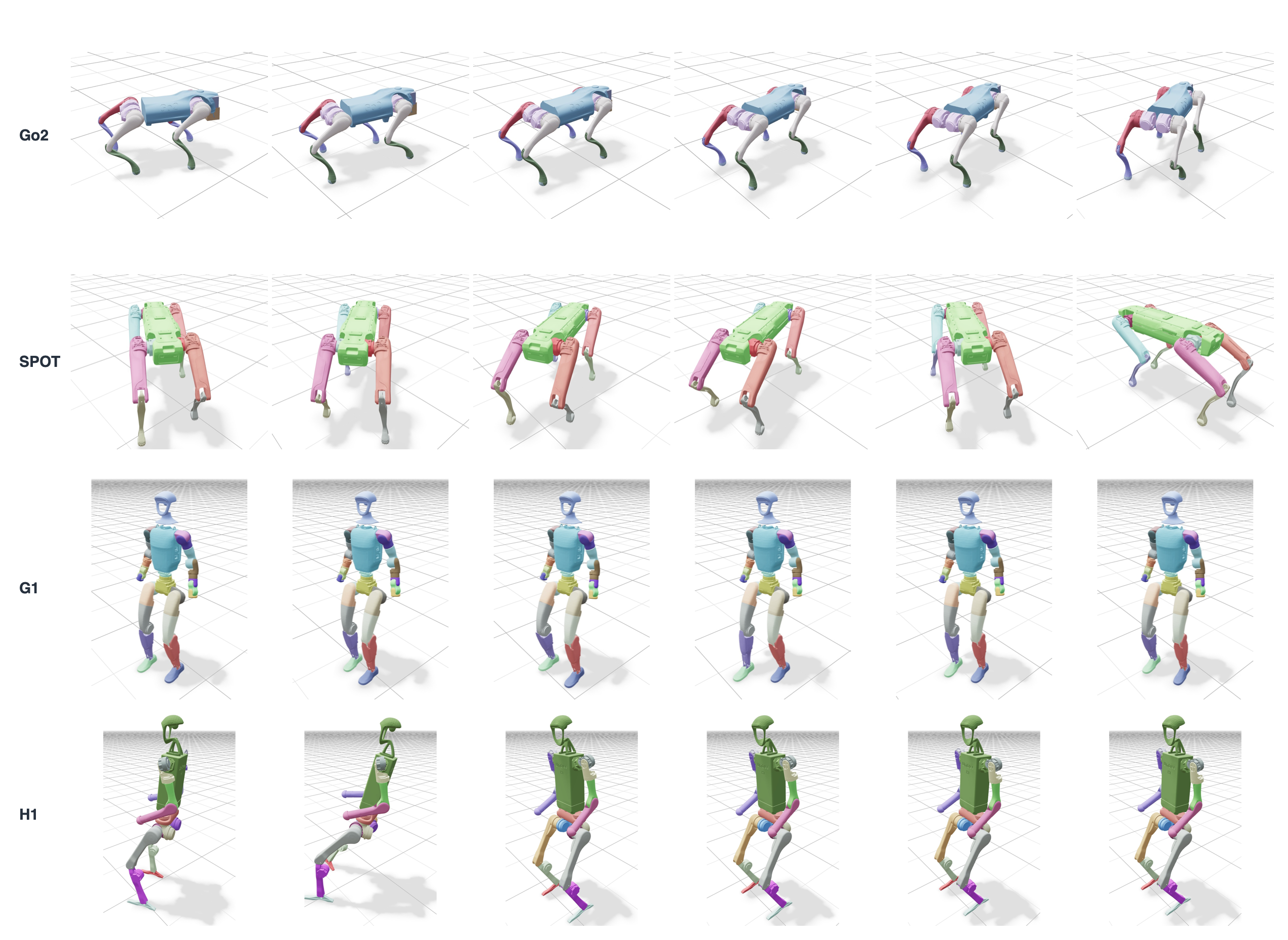}
\caption{\textbf{Cross-embodiment locomotion visualization (III).}
Representative simulation snapshots providing further qualitative visualizations of the learned locomotion policies.}
\label{fig:app_robot_rollouts_turning}
\end{figure*}

Figs.~\ref{fig:app_robot_rollouts_straight}--\ref{fig:app_robot_rollouts_turning} provide additional qualitative visualizations of simulated locomotion rollouts across Go2, Spot, G1, and H1.
The figures present representative snapshots of policy execution at different timestamps, illustrating locomotion behaviors across robotic embodiments with distinct morphologies and action spaces.
These visualizations complement the quantitative results reported in the main paper by providing qualitative observations of the learned one-step generative policies.

